# When ChatGPT is gone: Creativity reverts and homogeneity persists

| Qinghan Liu | Yiyong Zhou | Jihao Huang | Guiquan Li[*] |
|---|---|---|---|
| Peking U., School of Psychological and Cognitive Sciences | Peking U., School of Psychological and Cognitive Sciences | Beijing Yuxin Technology Company | Peking U., School of Psychological and Cognitive Sciences |

* Corresponding authors: liguiquan@pku.edu.cn

## ABSTRACT

ChatGPT has been evidenced to enhance human performance in creative tasks. Yet, it is still unclear if this boosting effect sustains with and without ChatGPT. In a pre-registered seven-day lab experiment and a follow-up survey after 30 days of experiment completion, we examined the impacts of ChatGPT presence and absence on sustained creativity using a text dataset of 3302 creative ideas and 427 creative solutions from 61 college students. Participants in the treatment group used ChatGPT in creative tasks, while those in the control group completed the tasks by themselves. The findings show that although the boosting effect of ChatGPT was consistently observed over a five-day creative journey, human creative performance reverted to baseline when ChatGPT was down on the 7th and the 30th day. More critically, the use of ChatGPT in creative tasks resulted in increasingly homogenized contents, and this homogenization effect persisted even when ChatGPT was absence. These findings pose a challenge to the prevailing argument that ChatGPT can enhance human creativity. In fact, generative AI like ChatGPT lends to human with a temporary rise in creative performance but boxes human creative capability in the long run, highlighting the imperative for cautious generative AI integration in creative endeavors.

## 1 Introduction

As noted in a Nature news feature from last December, "the AI system was a force in 2023 — for good and bad". Emblematic of this revolution is ChatGPT, a tool that has rapidly and profoundly infiltrated our lives, transforming everything from the labor market to the landscape of scientific publishing (Bertolo & Antonelli, 2023; Ghassemi et al., 2023; Kaiser, 2023; Nature Cancer , 2023; Noy & Zhang, 2023). A global survey by Nature, involving 1600 researchers, underlines this paradigm shift, revealing that a majority view AI tools as 'very important' or 'essential' for their research in the coming decade. Recognizing ChatGPT's extensive impact across various scientific disciplines, Nature has acknowledged it as one of the most influential entities in science for the year 2023 (Noorden & Webb, 2023.; Van Noorden & Perkel, 2023).

By harnessing vast amounts of human-generated data, such as the Internet, ChatGPT has exhibited remarkable capabilities in activities ranging from coding to creative writing, often surpassing human performance. Empirical evidence suggests that ChatGPT is instrumental in reducing labor market inequalities, enabling workers of varied skills to enhance productivity significantly (Noy & Zhang, 2023). Remarkably, in realms traditionally considered exclusive to human ingenuity—like literature, music, and art—generative AI tools have demonstrated extraordinary prowess (Nature Machine Intelligence, 2022; Epstein et al., 2023; Rafner, Beaty, Kaufman, Lubart, & Sherson, 2023). ChatGPT, for instance, has not only matched human creativity in various tests but has also been instrumental in amplifying it, heralding a new era of human-AI co-creativity (Folk, 2023; Jia, Luo, Fang, & Liao, 2023; Koivisto & Grassini, 2023; Rafner et al., 2023).

However, this technological marvel is not without its detractors. Concerns range from exacerbating biases and privacy breaches to perpetuating misinformation (Acion, Rajngewerc, Randall, & Etcheverry, 2023; Bockting, Van Dis, Van Rooij, Zuidema, & Bollen, 2023; Choudhury, 2023; Kidd & Birhane, 2023; Porsdam Mann et al., 2023). A particularly salient issue is the potential for content homogenization, as researchers and creators increasingly rely on AI tools trained on uniform databases (Nakadai, Nakawake, & Shibasaki, 2023; Wong & C, 2023). While this may initially facilitate information dissemination and spark imagination, it could, over time, hinder the diversity of thought crucial for groundbreaking innovations. This paradox places researchers at a crossroads, exhilarated by the transformative capabilities of tools like ChatGPT, yet cautious of their far-reaching implications.

## 1.1 Research questions and significance

As generative AI continues to evolve, the extent to which we should allow it to permeate our scientific work, and the degree to which we should permit opaque algorithms to replace our creative activities, are pressing questions needing answers (Acion et al., 2023; Jo & A, 2023; Nakadai et al., 2023). To address these, we need to delve deeper into three key questions. Research Question1: While using ChatGPT for creative activities may seem to enhance our performance in the short term, what are the long-term impacts of its use in creative work? Research Question2: What happens to our creative behavior if we become overly reliant on ChatGPT for co-creativity and then suddenly lose access to it? In other words, is the creativity augmented by ChatGPT genuine? Research Question3: Does ChatGPT hinder innovation, leading to a homogenization of ideas and possibly resulting in a loss of scientific diversity?

The advent of generative AI, akin to opening Pandora's Box, presents a complex landscape of opportunities and challenges. While AI assistance offers manifold benefits, it also represents a double-edged sword, necessitating a careful balance between leveraging its advantages and mitigating its risks.

To investigate the three research questions, this study enrolled 61 university students, randomly allocating them into an independent group or a ChatGPT-assisted group, for a comprehensive seven-day lab experiment followed by a one-month follow-up survey. Over the course of the study, they engaged in 16 diverse creative tasks, generating a robust dataset of 3302 ideas and 427 solutions. Specifically, participants were tasked with completing two types of creative assignments each day, either independently or with the aid of ChatGPT. The first task was a low-complexity alternative uses test (AUT) with a three-minute time limit, and the second, a high-complexity task involving the addition of new features to a product for a company, requiring solution proposals. This long-term laboratory study aimed to explore the impact of prolonged use of ChatGPT on individual continuous innovation capabilities, compared to completing creative tasks independently. We focused particularly on how people's innovative performances would change if they suddenly lost access to ChatGPT, having become overly dependent on it for creative tasks. Compared to the independent completion group, does the use of ChatGPT lead to a convergence in people's overall ideas, and does this convergence persist even after the removal of ChatGPT? In essence, we sought to investigate whether prolonged use of ChatGPT enhances creativity but leads to more homogenized content, and whether such enhancements in creativity and homogenization persist in future creative activities after ChatGPT is no longer used.

## 2 Method: One-week-experimental design

### 2.1 Participants

Sixty-one college students ($M_{age}$ = 21.56 years, $SD$ = 2.62) with 31 different majors from a university completed the lab experiment (per-registered at https://osf.io/hea8r). 36 of them were female. Participants were invited to complete a series of lab experiment that last for a week without intervals. After consenting to the study, they were randomly assigned to one of two conditions: ChatGPT use (the treatment group, $n$ = 31) and no ChatGPT use (the control group, $n$ = 30).

### 2.2 Materials and procedure

To capture the full picture of creativity, we adopted 14 different creative tasks to measure participant's divergent thinking and convergent thinking, two critical yet distinct pathway to creativity. Whereas divergent thinking involves the generation of multiple ideas in diverse directions (e.g., listing the unusual uses for a pen, Guilford, 1967), convergent thinking involves identifying the best solution to a clearly defined problem (e.g., Duncker's candle problem; Duncker, 1945).

All participants were asked to complete two types of creativity tasks each day: a low-complexity task (AUT: generating alternative uses for a product) and a high-complexity task (Come up with innovative function points and propose solutions for the products developed in the enterprise). In the first AUT task, participants had a maximum of 3 minutes to list creative uses for everyday objects, such as a pen or a brick. Then, the experimenter would ask the participants to find the best way to solve a real-life problem (Task 2), such as developing new functions for a VR glasses. For task 2, there is no time limited. The orders of tasks were counter-balanced.

In the first day and the last day, all participants were invited to finish the same two creative tasks without any ChatGPT assistance. We did so to make sure there was no significant difference on baseline creativity (Day1). More importantly, we were curious about whether there was a remaining manipulated impacts of ChatGPT use experiences on creativity performance (Day7).

We manipulated the ChatGPT use experiences during the middle 5 days (from Day 2 to Day 6). We provided half of the participants (the treatment group) ChatGPT 4.0 to help them to finish the creative tasks, while the other half still finished the same two creative tasks by themselves (the control group). The flow of experimental design was depicted in Figure 1.

**Figure1 Experimental Design**

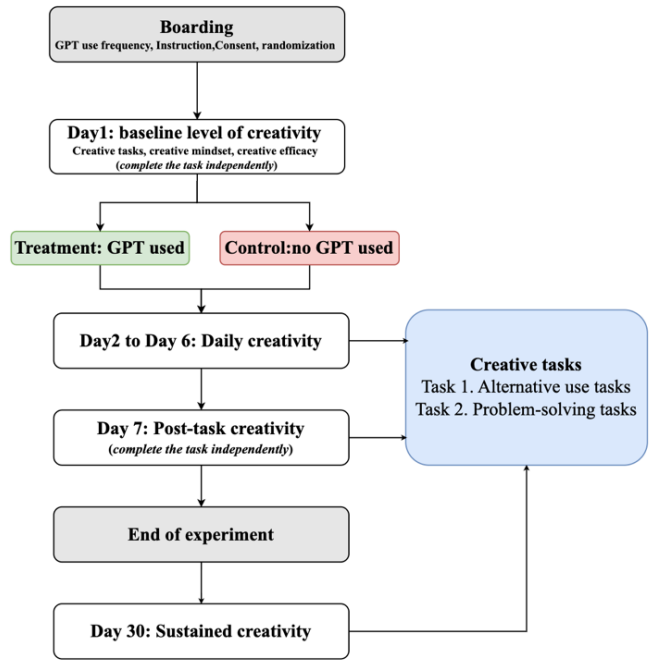

## 2.3 Creativity code and assessment

Two experimenters were trained to code the text content. For AUT task, we extracted independent ideas from the text content. Preliminary analysis showed that participants came up with at least 351 (Day 7) to maximum 501 (Day 4) ideas per day, averaged at 5.75 ideas per person.

For problem-solving task, we remained all their answers. We randomized the order in a spreadsheet without any identifying information.

We had four (two for each task) coders who were blind to the study purposes and experimental conditions rate participants' responses, adherent to the consensual assessment technique (CAT, Amabile, 1996). We provided raters with the spreadsheet and instructed them to evaluate each idea in it.

*Assessment on divergent thinking.* In AUT tasks, two raters were instructed to independently rate participants' each idea in terms of novelty (i.e., the originality of the idea) and usefulness (i.e., the practicality of the idea). The interrater reliabilities for both aspects were satisfactory (ICCs ≥ .90). Participants' level of novelty and usefulness was respectively computed by summing the scores of each idea. Flexibility of each participant was subjective and was obtained by counting the total number of unique usage categories. Besides, we asked participants to select the most creative idea when they complected the AUT task. So, we computed participants' self-recognition accuracy by subtracting the score of the selected idea from the maximum score of this participant.

*Assessment on convergent thinking.* In problem-solving task, another two raters who had rich work experiences on new project development were instructed to independently rate participants' responses in four different ways: (1) creativity (i.e., the novel and useful level of idea); (2) content quality (i.e., the writing, structures, and logics of the idea), (3) public popularity (i.e., prediction on customer's favor if the idea was implemented), (4) market success (i.e., prediction on product success if the idea was implemented). The interrater reliabilities for all aspects were satisfactory (ICCs ≥ .90).

## 3 Results

### 3.1 The impact of ChatGPT use on creativity performance

*Baseline test*. Based on the results of Day 1, there was no significant differences on creativity performance between two groups, implying that all participants' creativity were similar.

***The manipulated effects of ChatGPT use on everyday creative tasks***. According to the results of Day 2 to Day 6, participants in the ChatGPT condition performed superior to those in the control condition. To be specific, participants with the help of ChatGPT generated more novel, useful, and flexible usages in the AUT tasks. What is more, when asked to solve a real-life problem, they also came up with more creative and highly valuable ideas with more potential to gain public favor and market success.

However, we found that ChatGPT failed to help participant to recognize the most creative idea in AUT task, as we observed no significant difference on novelty and usefulness recognition accuracy between two groups.

***The effects of ChatGPT on sustained creativity.*** Based on the results of Day 7, in which all participants completed the creative tasks without any ChatGPT assistance, participants who used ChatGPT in the last five days showed a sharp decrease in both divergent thinking and convergent thinking. Specifically, they did not perform better in all kinks of creativity, such as novelty, usefulness, flexibility, content quality, possible public favors, and market success, compared to the participants in control group (all p-values were insignificant). One month later (label as Day 30), in the follow-up survey where participants were invited to complete another AUT task (i.e., sponge), we repeated the findings that there was no significant difference on creativity in terms of novelty, usefulness, and flexibility between participants from the treatment versus control group. Taken together, we concluded that although ChatGPT helps people excel in creative tasks, people's creativity could not sustain but fall back to the average level once they did not have ChatGPT to rely on.

We utilized the R packages broom and dplyr for conducting t-tests to compare differences between the treatment and control groups. For data visualization, the ggplot2 package was employed. The results of creative scores are presented in Table 1 and Table 2. The visualization of these results can be found in Figures 2a and 2b (Only parts related to novelty and usefulness in Figure 2a and 2b are shown; see appendix for the rest..).

**Table2a Results of T-test on divergent thinking (Task 1)**

|       | novelty |       | usefulness |       | flexibility |       | novelty recognition accuracy |       | usefulness recognition accuracy |       |
|-------|---------|-------|------------|-------|-------------|-------|------------------------------|-------|---------------------------------|-------|
|       | Control | GPT   | Control    | GPT   | Control     | GPT   | Control                      | GPT   | Control                         | GPT   |
| Day 1 | 7.50    | 8.78  | 29.92      | 34.35 | 5.87        | 6.77  | 0.53                         | 0.52  | 2.08                            | 1.95  |
| Day 2 | 30.77   | 41.55 | 19.22      | 30.58 | 6.30        | 8.16  | 3.13                         | 2.63  | 2.00                            | 2.10  |
| Day 3 | 28.37   | 40.58 | 16.32      | 28.98 | 4.58        | 6.24  | 3.31                         | 2.27  | 1.40                            | 1.27  |
| Day 4 | 17.17   | 35.32 | 29.98      | 52.63 | 6.10        | 9.81  | 1.52                         | 1.42  | 0.88                            | 1.19  |
| Day 5 | 18.17   | 41.44 | 26.10      | 49.44 | 5.93        | 9.10  | 1.33                         | 1.11  | 1.58                            | 1.10  |
| Day 6 | 14.80   | 39.84 | 28.67      | 51.87 | 5.60        | 8.71  | 1.23                         | 1.05  | 1.28                            | 0.66  |
| Day 7 | 13.02   | 17.60 | 27.27      | 26.73 | 5.80        | 5.71  | 0.97                         | 1.03  | 1.68                            | 1.95  |
| Day30 | 2.40    | 2.56  | 5.62       | 5.55  | 4.27        | 5.39  | /                            | /     | /                               | /     |

*Note*: Numbers in bold are not significant, while the rests are significant (p ≤ .001)

**Table 2b Results of T-test on convergent thinking (Task 2)**

|       | Creativity |       | writing quality |       | popularity |       | Success |       |
|-------|------------|-------|-----------------|-------|------------|-------|---------|-------|
|       | Control    | GPT   | Control         | GPT   | Control    | GPT   | Control | GPT   |
| Day 1 | **3.97**   | **4.35** | **3.87**     | **4.02** | **4.42** | **4.63** | **3.73** | **3.79** |
| Day 2 | 2.83       | 3.82  | 3.13            | 4.60  | 3.27       | 4.55  | 2.15    | 3.55  |
| Day 3 | 3.10       | 3.92  | 3.53            | 4.55  | 3.50       | 4.44  | 2.68    | 3.60  |
| Day 4 | 2.85       | 3.29  | 2.95            | 4.03  | 3.28       | 4.27  | 2.23    | 3.39  |
| Day 5 | 3.07       | 3.81  | 3.32            | 4.32  | 3.57       | 4.53  | 2.53    | 3.53  |
| Day 6 | 2.95       | 3.76  | 3.42            | 4.73  | 2.90       | 4.34  | 2.45    | 3.66  |
| Day 7 | **3.38**   | **3.90** | **3.15**     | **3.56** | **3.62** | **3.87** | **2.92** | **3.21** |

*Note*: Numbers in bold are not significant, while the rests are significant (p ≤ .001)

**Figure2a Task 1 (AUT) Bar chart of each variable daily**

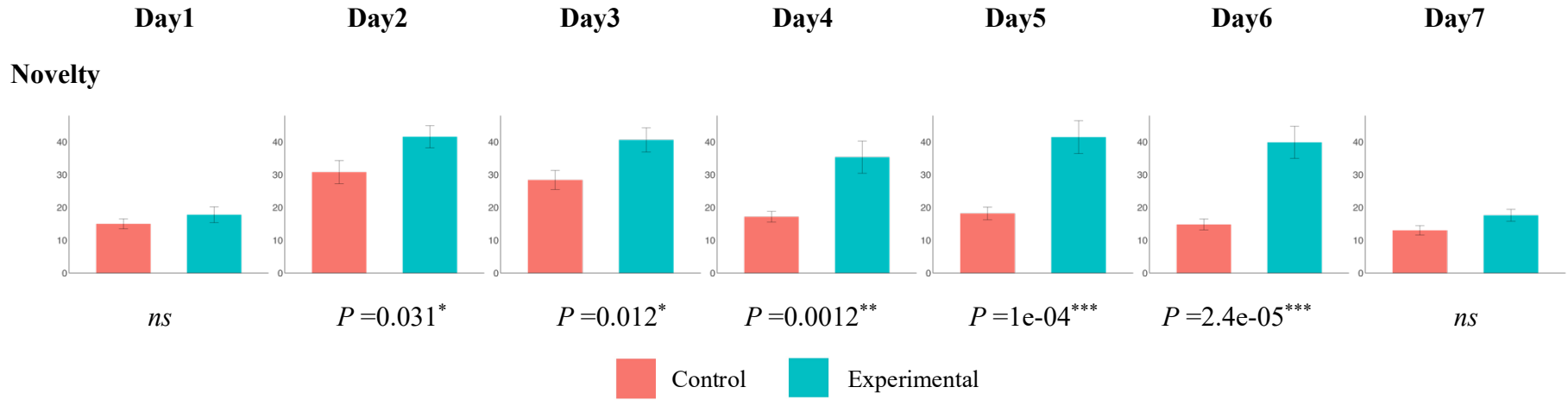

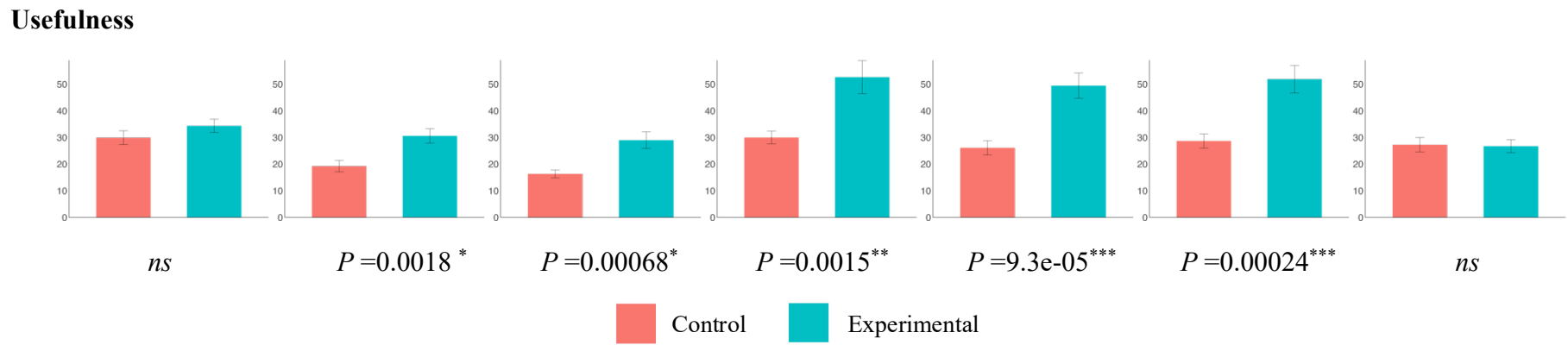

**Figure2b Task 2 Bar chart of each variable daily**

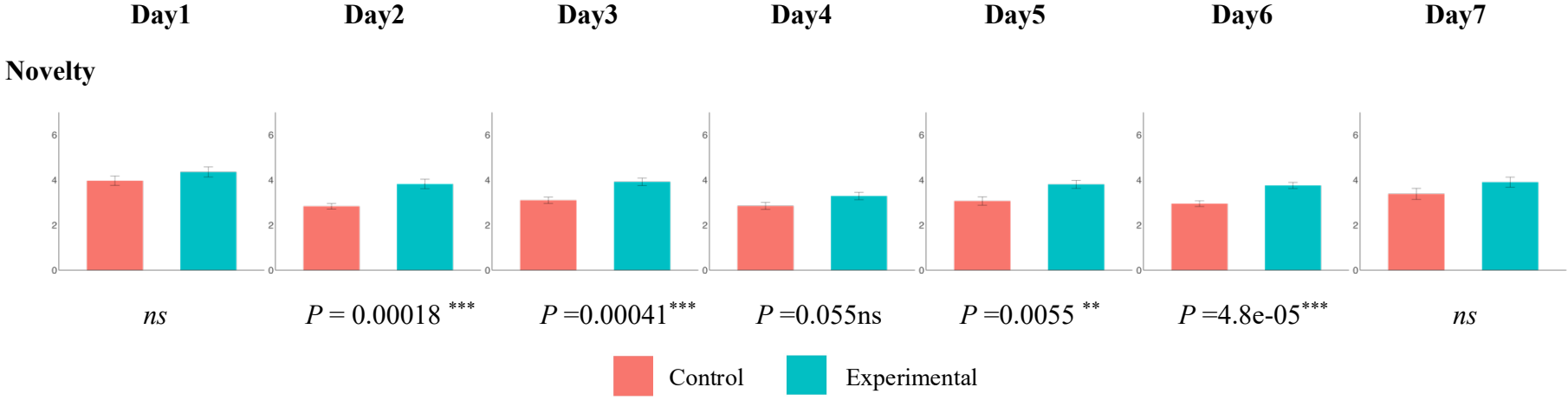

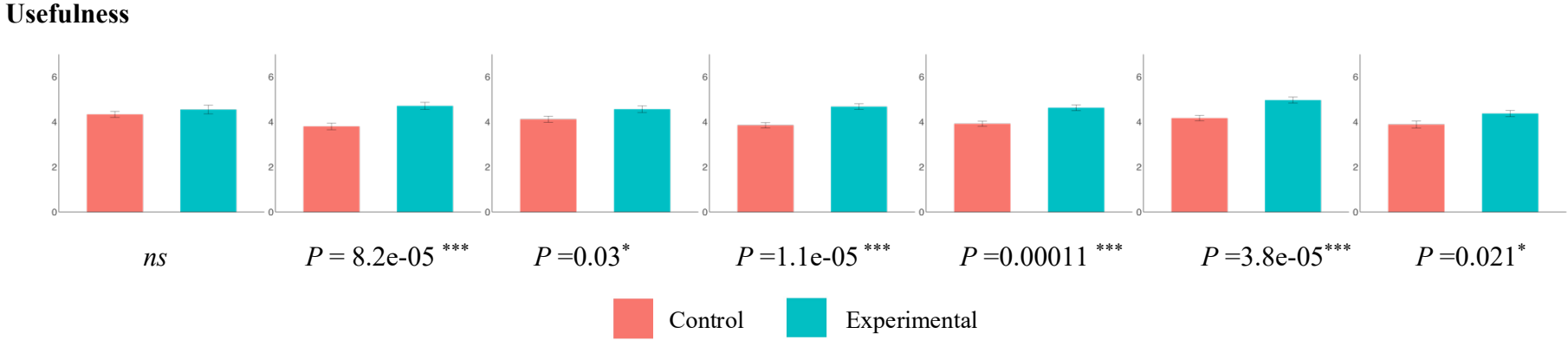

**Table1 T-Test Results for Daily Variables**

| Day | Task1-Variable | Estimate | Statistic | P Value | Significance | Day | Task2-Variable | Estimate | Statistic | P Value | Significance |
|---|---|---|---|---|---|---|---|---|---|---|---|
| 1 | novelty | -1.28 | -0.90 | 0.373 | ns | 1 | novelty | -0.39 | -1.28 | 0.206 | ns |
| 1 | usefulness | -4.44 | -1.23 | 0.222 | ns | 1 | usefulness | -0.22 | -0.92 | 0.362 | ns |
| 1 | Judgement N | 0.01 | 0.05 | 0.961 | ns | 1 | content quality | -0.15 | -0.62 | 0.535 | ns |
| 1 | Judgement U | 0.13 | 0.32 | 0.751 | ns | 1 | public favor | -0.21 | -0.96 | 0.340 | ns |
| 2 | novelty | -10.78 | -2.21 | 0.031 | * | 1 | market success | -0.06 | -0.22 | 0.829 | ns |
| 2 | usefulness | -11.36 | -3.28 | 0.002 | ** | 2 | novelty | -0.99 | -4.05 | 0.000 | *** |
| 2 | Judgement N | 0.50 | 1.06 | 0.292 | ns | 2 | usefulness | -0.91 | -4.23 | 0.000 | *** |
| 2 | Judgement U | -0.10 | -0.15 | 0.882 | ns | 2 | content quality | -1.46 | -5.21 | 0.000 | *** |
| 3 | novelty | -12.21 | -2.60 | 0.012 | * | 2 | public favor | -1.28 | -5.35 | 0.000 | *** |
| 3 | usefulness | -12.67 | -3.66 | 0.001 | ** | 2 | market success | -1.40 | -5.47 | 0.000 | *** |
| 3 | Judgement N | 1.03 | 2.67 | 0.010 | * | 3 | novelty | -0.82 | -3.75 | 0.000 | *** |
| 3 | Judgement U | 0.13 | 0.48 | 0.635 | ns | 3 | usefulness | -0.45 | -2.22 | 0.030 | * |
| 4 | novelty | -18.16 | -3.51 | 0.001 | ** | 3 | content quality | -1.02 | -4.06 | 0.000 | *** |
| 4 | usefulness | -22.65 | -3.41 | 0.002 | ** | 3 | public favor | -0.94 | -3.89 | 0.000 | *** |
| 4 | Judgement N | 0.10 | 0.24 | 0.814 | ns | 3 | market success | -0.91 | -3.90 | 0.000 | *** |
| 4 | Judgement U | -0.31 | -1.23 | 0.224 | ns | 4 | novelty | -0.44 | -1.96 | 0.055 | ns |
| 5 | novelty | -23.27 | -4.34 | 0.000 | *** | 4 | usefulness | -0.83 | -4.80 | 0.000 | *** |
| 5 | usefulness | -23.34 | -4.27 | 0.000 | *** | 4 | content quality | -1.08 | -4.76 | 0.000 | *** |
| 5 | Judgement N | 0.22 | 0.66 | 0.509 | ns | 4 | public favor | -0.99 | -4.57 | 0.000 | *** |
| 5 | Judgement U | 0.49 | 1.45 | 0.154 | ns | 4 | market success | -1.15 | -5.21 | 0.000 | *** |
| 6 | novelty | -25.04 | -4.84 | 0.000 | *** | 5 | novelty | -0.74 | -2.88 | 0.006 | ** |
| 6 | usefulness | -23.20 | -3.99 | 0.000 | *** | 5 | usefulness | -0.71 | -4.16 | 0.000 | *** |
| 6 | Judgement N | 0.18 | 0.44 | 0.660 | ns | 5 | content quality | -1.01 | -4.11 | 0.000 | *** |
| 6 | Judgement U | 0.62 | 2.76 | 0.008 | ** | 5 | public favor | -0.97 | -3.69 | 0.000 | *** |
| 7 | novelty | -4.58 | -1.98 | 0.053 | ns | 5 | market success | -1.00 | -4.51 | 0.000 | *** |

| | | | | | | | | | | |
|---|---|---|---|---|---|---|---|---|---|---|
| 7 | usefulness | 0.54 | 0.15 | 0.882 | ns | 6 | novelty | -0.81 | -4.39 | 0.000 | *** |
| 7 | Judgement N | -0.07 | -0.20 | 0.842 | ns | 6 | usefulness | -0.80 | -4.46 | 0.000 | *** |
| 7 | Judgement U | -0.27 | -0.67 | 0.504 | ns | 6 | content quality | -1.31 | -4.68 | 0.000 | *** |
| 30 | novelty | -3.69 | -2.12 | 0.038 | * | 6 | public favor | -1.44 | -5.76 | 0.000 | *** |
| 30 | usefulness | -6.26 | -1.52 | 0.135 | ns | 6 | market success | -1.21 | -5.57 | 0.000 | *** |
| | | | | | | 7 | novelty | -0.52 | -1.57 | 0.122 | ns |
| | | | | | | 7 | usefulness | -0.49 | -2.37 | 0.021 | * |
| | | | | | | 7 | content quality | -0.41 | -1.54 | 0.130 | ns |
| | | | | | | 7 | public favor | -0.25 | -0.87 | 0.387 | ns |
| | | | | | | 7 | market success | -0.29 | -1.03 | 0.308 | ns |

*Note*: N=61. Day denotes specific days; Task1-Variable pertains to AUT task; Task2-Variables relates to problem-solving tasks; Judgement N representing the accuracy of identifying of novel ideas; Judgement U representing the accuracy of identifying useful ideas ; ns=not significant.* $p < .05$; ** $p < .01$; *** $p < .001$

## 3.2 The impacts of ChatGPT dependence on knowledge homogeneity

We examined whether ChatGPT use caused knowledge homogeneity. In another words, we were interested to know whether participant with the help from ChatGPT generated more creative but less diverse responses. If it is true, we would further explore whether this impact would remain even they did not use ChatGPT in the future creative tasks (i.e., Day 7 and Day 30 in this study). We mainly relied on Machine Learning and the Natural Language Processing technology to estimate the homogeneity of participants' responses.

***Using SBERT for text semantic similarity calculation.*** Sentence-BERT (Sentence-Bidirectional Encoder Representation from Transformers, shorten as SBERT) is an improved model based on BERT, designed to better handle sentence-level embeddings (Reimers & Gurevych, 2019). Unlike the original BERT model, which encodes individual words, SBERT encodes entire sentences, providing a richer semantic representation. This feature gives SBERT a significant advantage in tasks requiring understanding of the entire sentence semantics, such as text similarity calculation.

To use SBERT for text semantic similarity, we first trained the SBERT model. During training, the model learns semantic relationships between a large number of sentences. Once trained, we used this model to encode sentences from creative texts, obtaining their semantic representations (sentence embedding). Then, we calculated the cosine similarity between these semantic representations of the creative texts to determine their similarity. For example, with two answer "S1" and "S2", we first encoded them using the SBERT model to obtain two sentences embedding "Embed1" and "Embed2". We then calculated the cosine similarity between "Embed1" and "Embed2" as the similarity between "S1" and "S2". The formula for cosine similarity is as follows:

$$\cos(\theta) = (A \cdot B) / (\|A\| \|B\|)$$

A and B are two sentences embedding, "·" denotes the dot product of the vectors, and "$\|A\|$" and "$\|B\|$" are the lengths (or norms) of vectors A and B, respectively. During the calculation, we first computed the dot product of sentence embedding A and B, then calculated the norms of sentence embedding A and B, and finally divided the dot product by the product of the norms, resulting in the cosine similarity.

The flow of SBERT was presented in Figure 3.

**Figure3**

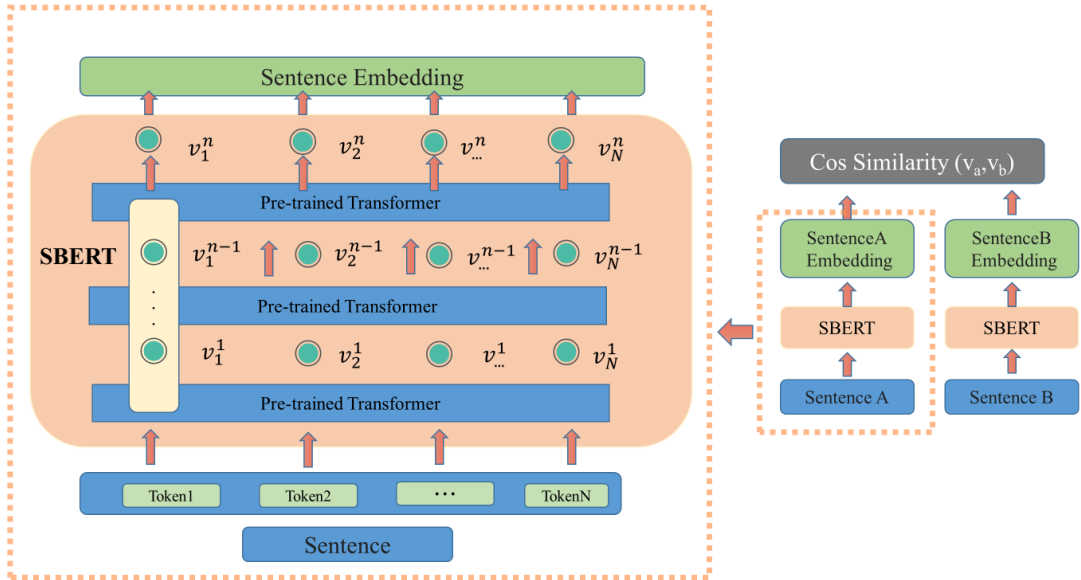

Based on the results of Day 1 using SBERT, we found that all participants shared a heterogeneous response towards to the tasks (the similarity of $M_C$ = 0.36, the similarity of $M_E$ = 0.35, *n.s*). However, the knowledge diversity became lost when some of them began to use ChatGPT to generate answers (the similarity of $M_C$ = [0.3, 0.48], the similarity of $M_E$ = [0.35, 59], *ps* ≤ .01), and it remained a decreasing trend in one month later (the similarity of $M_C$ = 0.33, the similarity of $M_E$ = 0.39, $p$ ≤ .001), in which they did not use ChatGPT to complete the creative tasks at all. As a comparison, we did not find the same pattern in participants who did not use ChatGPT during the whole study. To cross-validate the findings, we also used String Matching algorithm to compute the common string in the context, another index to reflect the context similarity.

*Using String Matching for text redundancy calculation.* String matching was implemented through a series of algorithms that can identify and compare common character sequences between two texts. We used string matching techniques to assess the redundancy between two creative texts. This method is widely used in academia to detect plagiarism in papers. Our goal was to determine the "literal" similarity between creative texts, which was achieved by identifying identical string sequences in two answers.

In this study, we employed an improved String Matching Technique that compares not only individual characters but also considers the sequence of characters in a string. Our algorithm first converted the two creative texts "S1" and "S2" into separate strings and then searched for repeated sequences longer than two characters. This method helped to exclude accidental or insignificant matches, thereby increasing the accuracy of redundancy assessment. The calculation process includes several key steps: 1) Preprocessing: Convert "S1" and "S2" into a standard format to ensure consistency. 2) Match Searching: Search for repeated strings longer than two characters in both sentences. 3) Redundancy Calculation: Calculate the mean of repeated strings for a task. We found that by considering strings longer than two characters, we can effectively assess the redundancy rate between the two creative texts "S1" and "S2", while avoiding overemphasis on short, common phrases. This method provided a more precise framework for understanding the substantive similarity between two sentences.

The results of Day 1 showed that there was a relatively low rate of common strings towards to the tasks (the redundancy of $M_C$ = 0.36, the redundancy of $M_E$ = 0.35, *n.s*). However, in the next five days when participants used ChatGPT to generate answers, their responses shared a larger rate of common strings than the participants who did not use the ChatGPT (the redundancy of $M_C$ = [3.13, 3.4], the redundancy of $M_E$ = [5.90, 8.73], *ps* ≤ .01). So, the knowledge homogeneity caused by ChatGPT use was repeated.

Table 3 Results of similarity and redundancy of the context

| TASK1 | SBERT Control | SBERT GPT | String Matching Control | String Matching GPT | TASK2 | SBERT Control | SBERT GPT | String Matching Control | String Matching GPT |
|---|---|---|---|---|---|---|---|---|---|
| Day 1 | 0.36 | 0.35 | 3.44 | 3.51 | Day 1 | 0.59 | 0.58 | 70.4 | 64.5 |
| Day 2 | 0.39 | 0.48 | 2.73 | 5.9 | Day 2 | 0.51 | 0.62 | 55.82 | 215.31 |
| Day 3 | 0.32 | 0.48 | 3.17 | 8.24 | Day 3 | 0.69 | 0.83 | 57.65 | 214.57 |
| Day 4 | 0.3 | 0.45 | 3.33 | 7.12 | Day 4 | 0.57 | 0.62 | 52.41 | 203.87 |
| Day 5 | 0.48 | 0.59 | 3.4 | 8.73 | Day 5 | 0.55 | 0.72 | 63.71 | 195.38 |
| Day 6 | 0.38 | 0.53 | 3.13 | 7.95 | Day 6 | 0.63 | 0.69 | 63.57 | 209.03 |
| Day 7 | 0.33 | 0.36 | 2.76 | 3.74 | Day 7 | 0.6 | 0.62 | 54.74 | 75.28 |
| Day30 | 0.33 | 0.39 | 1.45 | 2.34 | Day30 | / | / | / | / |

Note: Numbers in bold are not significant, while the rests are significant ($p \leq .001$).

To visualize the above findings, we utilized the SBERT technology to generate sentences semantic embedding representations for the creativity responses of Group C (the control group) and Group E (the experimental group), so that we could delved into the semantic content of the answers. To efficiently reduce dimensions and visualize these complex data sets, we initially applied the PCA algorithm with parameters set to n_components = 30. This step was instrumental in diminishing the data dimensions while preserving essential information. Subsequently, we employed the UMAP algorithm with parameters including n_components = 2, n_neighbors = 5, min_dist = 0.001, spread = 1.0, metric= 'cosine', init = 'spectral', random_state = 0, and n_jobs = 1, to further decrease dimensions, achieving a more refined two-dimensional representation. The parameter settings for UMAP were specifically chosen to optimize data layout, facilitating clearer differentiation between distinct data points. Finally, by normalizing the data within the range of 0 to 1 using MinMaxScaler, we ensured the uniformity and comparability of the data. Through this series of processing steps, we successfully created semantic embedding density distribution maps for the creativity responses of Groups C and E, thereby visually depicting the differences in creativity between the two groups. The covariance matrix for each group was represented by an ellipse, with the ellipse's major and minor axes corresponding to the eigenvalues of the covariance matrix, and its orientation corresponding to the eigenvectors.

Once again, the figures demonstrated that the responses generated by the assistance ChatGPT (versus by human-only) generally exerted less standard deviation, eigenvalues, and ellipse area. What is more, this tendency was continuing at Day 7 and Day 30.

**Table 4a Results of density (Task 1)**

|  | Standard Deviation | | Sum of Eigenvalues | | Ellipse Area | |
|---|---|---|---|---|---|---|
|  | Control | GPT | Control | GPT | Control | GPT |
| Day1 | 0.34 | 0.32 | 0.12 | 0.10 | 0.17 | 0.15 |
| Day2 | 0.40 | 0.33 | 0.16 | 0.11 | 0.24 | 0.14 |
| Day3 | 0.34 | 0.41 | 0.12 | 0.17 | 0.17 | 0.14 |
| Day4 | 0.37 | 0.32 | 0.14 | 0.10 | 0.22 | 0.16 |
| Day5 | 0.36 | 0.33 | 0.13 | 0.11 | 0.19 | 0.17 |
| Day6 | 0.37 | 0.32 | 0.14 | 0.10 | 0.21 | 0.14 |
| Day7 | 0.37 | 0.38 | 0.14 | 0.15 | 0.22 | 0.22 |
| Day30 | 0.39 | 0.32 | 0.15 | 0.10 | 0.23 | 0.15 |

**Table 4b Results of density (Task 2)**

|  | Standard Deviation | | Sum of Eigenvalues | | Ellipse Area | |
|---|---|---|---|---|---|---|
|  | Control | GPT | Control | GPT | Control | GPT |
| Day1 | 0.45 | 0.50 | 0.21 | 0.25 | 0.31 | 0.15 |
| Day2 | 0.48 | 0.49 | 0.23 | 0.24 | 0.22 | 0.20 |
| Day3 | 0.46 | 0.53 | 0.21 | 0.28 | 0.33 | 0.14 |
| Day4 | 0.48 | 0.53 | 0.23 | 0.28 | 0.35 | 0.15 |
| Day5 | 0.51 | 0.54 | 0.26 | 0.29 | 0.29 | 0.18 |
| Day6 | 0.47 | 0.53 | 0.22 | 0.28 | 0.32 | 0.39 |
| Day7 | 0.50 | 0.58 | 0.25 | 0.34 | 0.13 | 0.12 |

Taken together, we consistently found that five-days of ChatGPT use led to serious homogeneity, in which participants generated more creative but less heterogeneous responses. What is worse, we found a fraught, continuing effect in knowledge homogeneity in future tasks, in which people generated AI-like responses even though they did no use ChatGPT.

**Figure4a Task1 (AUT): Comparison of the daily density distribution between the experimental group and the control group**

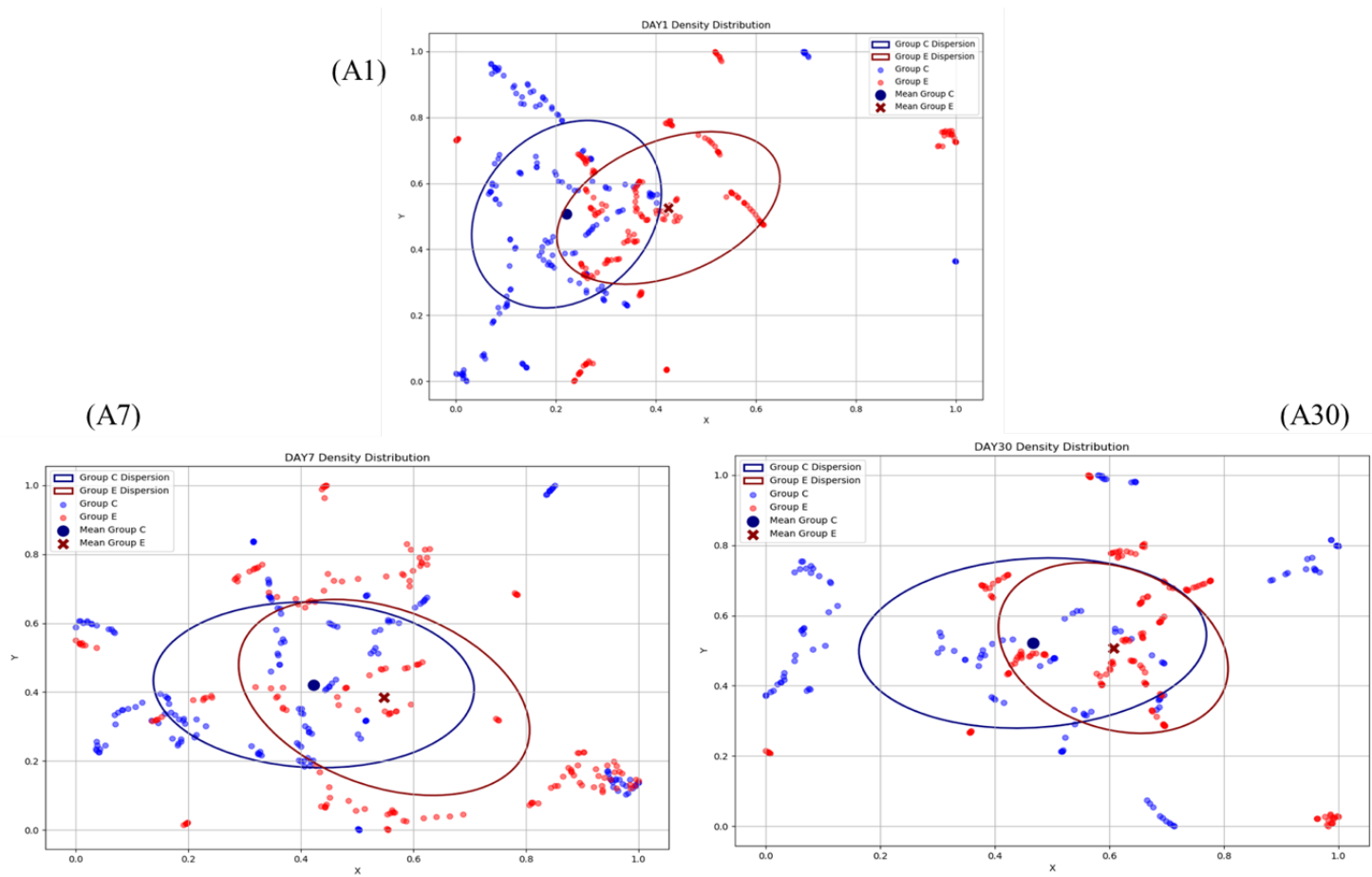

*Note*: 'A' presents Task1. The number following 'A' indicates the corresponding day of data (for example, 'A1' represents Day 1, 'A2' represents Day 2). Blue dots represent the control group, red dots represent the experimental group.

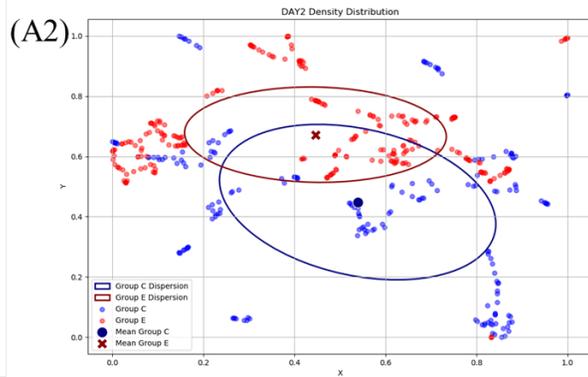
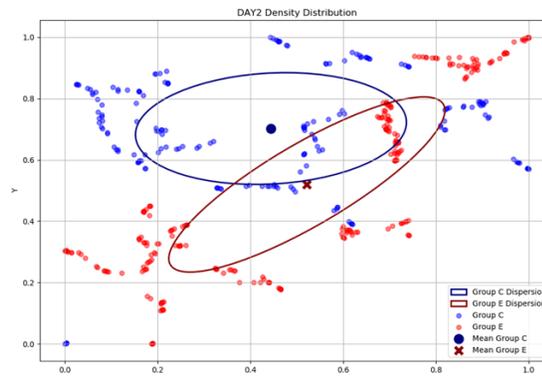
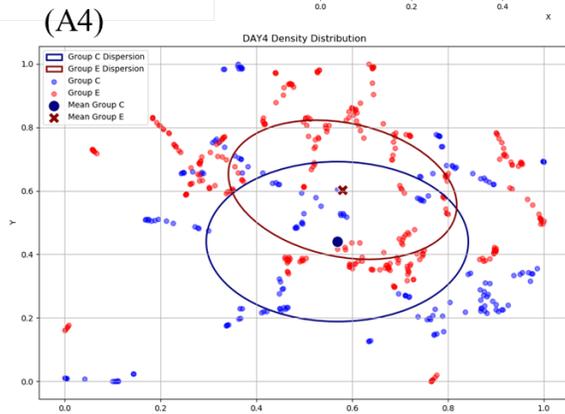
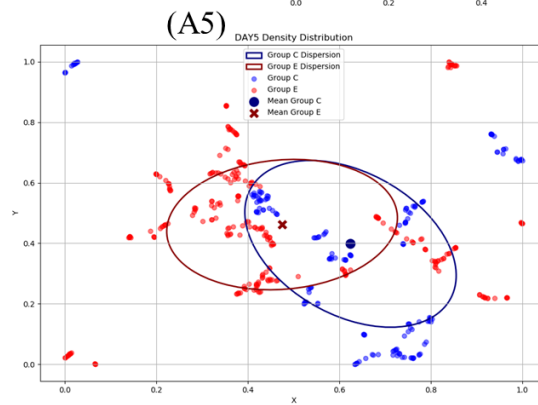
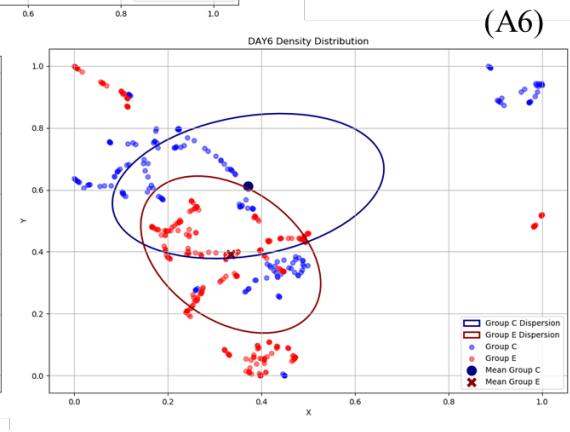

*Note*: 'A' presents Task1. The number following 'A' indicates the corresponding day of data (for example, 'A1' represents Day 1, 'A2' represents Day 2). Blue dots represent the control group, red dots represent the experimental group.

**Figure4b Task2: Comparison of the daily density distribution between the experimental group and the control group**

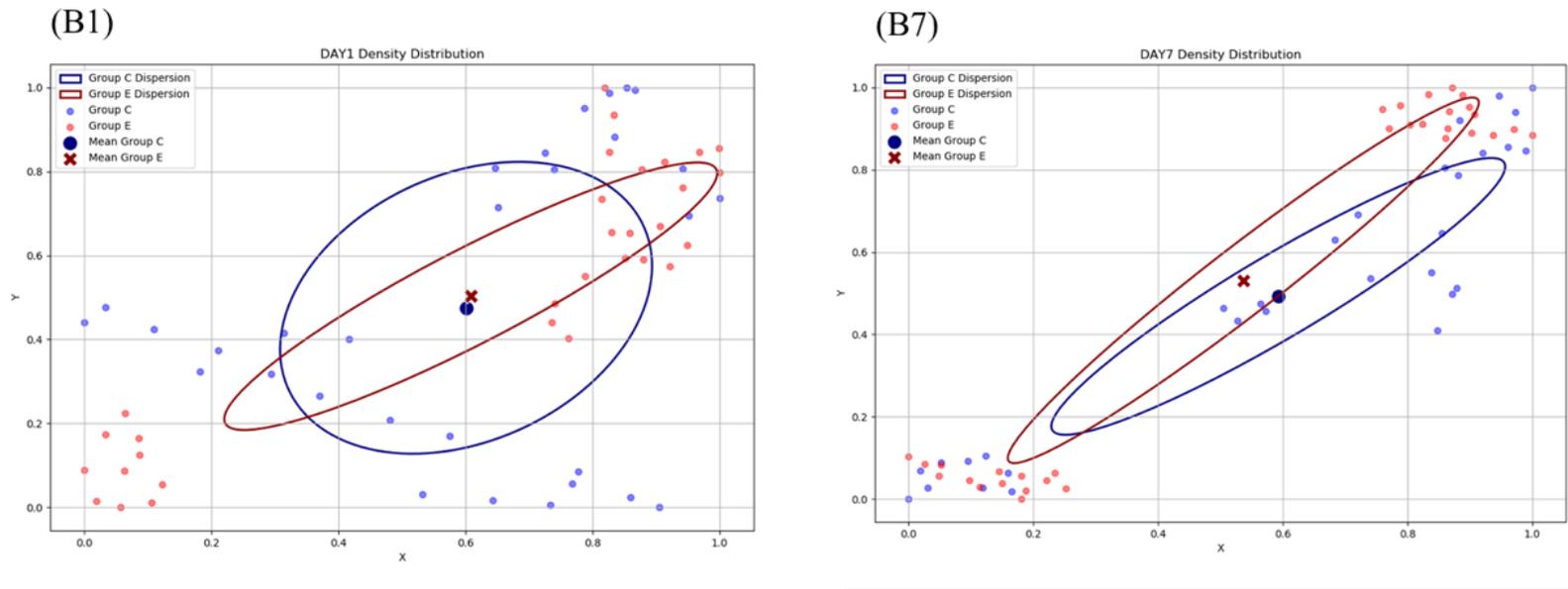

*Note*: 'B' presents Task2. The number following 'B' indicates the corresponding day of data (for example, 'B1' represents Day 1, 'B2' represents Day 2). Blue dots represent the control group, red dots represent the experimental group.

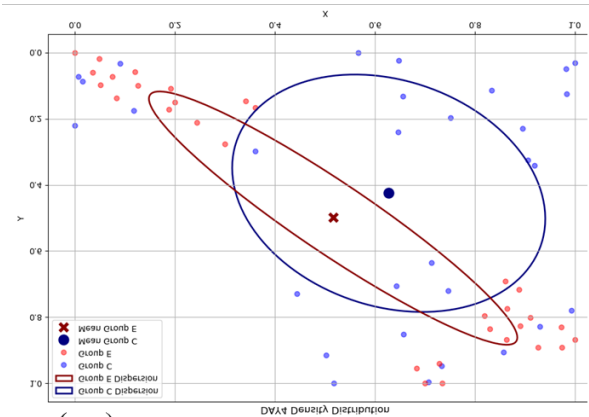
(B4)

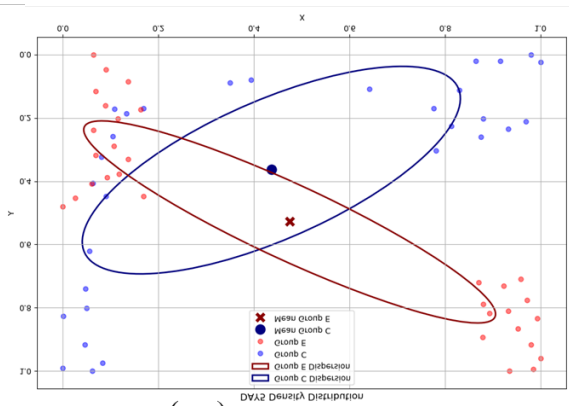
(B5)

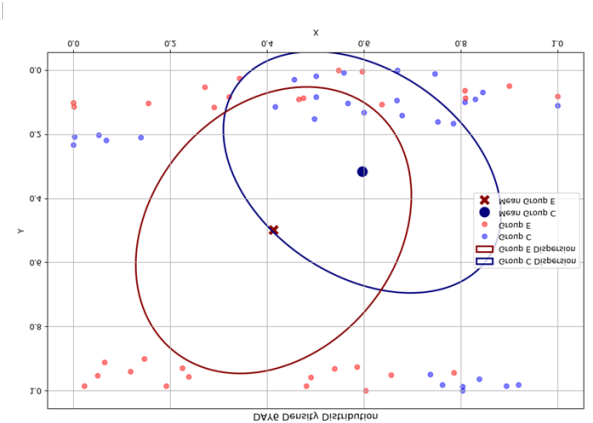
(B6)

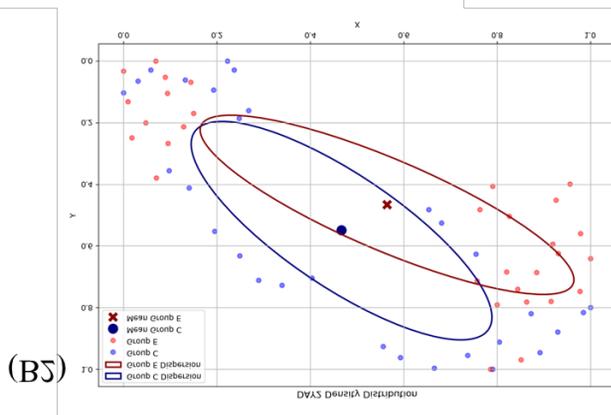
(B2)

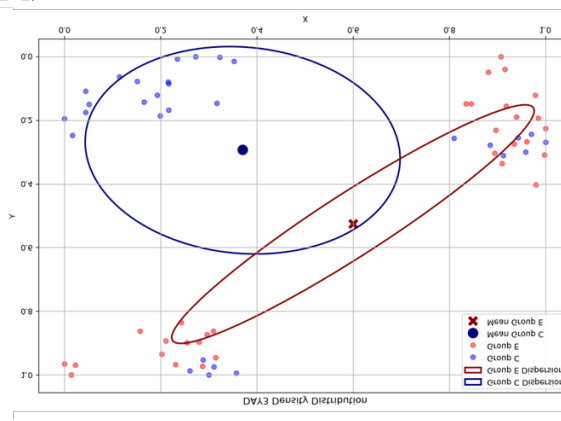
(B3)

*Note*: 'B' presents Task2. The number following 'B' indicates the corresponding day of data (for example, 'B1' represents Day 1, 'B2' represents Day 2). Blue dots represent the control group, red dots represent the experimental group.

# 4 Discussion

Individuals with formal university education, when engaging in Co-creativity using ChatGPT, exhibit a rapid enhancement in creativity levels in the short term, regardless of whether they are working on low-complexity AUT tasks or high-complexity tasks such as devising new features for corporate products. These findings are consistent with other studies in the field. However, our research goes a step further by uncovering that the group assisted by ChatGPT (the experimental group) generated ideas with significantly higher homogeneity than the group working independently (the control group). This indicates that the assistance of ChatGPT, due to the uniformity of data sources and tools used, leads to a convergence of ideas at a group level. Consequently, the use of ChatGPT appears to constrain the diversity of innovation. Even more striking is the long-term impact: 30 days later, upon reassessing both groups with an AUT task completed independently, we observed that the experimental group, devoid of ChatGPT's assistance, regressed to average creativity levels, aligning with the control group. However, the homogeneity of ideas in the experimental group remained higher. This suggests that the boost in individual creativity afforded by ChatGPT is transient. When individuals revert to independent creative activities without ChatGPT, their creativity capabilities diminish; moreover, from a group perspective, long-term reliance on ChatGPT for creative tasks ironically increases homogeneity of thought, significantly reducing the diversity of group ideas. This effect remains pronounced even after one month.

From a broader societal standpoint, unrestricted use of generative AI tools like ChatGPT in scientific research, brainstorming, and innovative activities could lead to unintended and potentially disruptive consequences for the scientific community (Nakadai et al., 2023). This unrestricted application could stifle the diversity of ideas, leading to increasingly similar thoughts among scientists worldwide, and a gradual diminution of unique innovations.

### 4.1 Implications

Firstly, our study advances the understanding of AI's role in enhancing creativity. The observed temporary boost in creativity, facilitated by ChatGPT, holds significant implications for the integration of AI in domains that are heavily reliant on creative processes. This insight is crucial for educational psychologists and creative professionals, underscoring AI's nuanced role in these fields. While AI tools like ChatGPT can significantly aid in immediately resolving creative challenges, their impact in nurturing long-term creative thinking skills seems limited. This finding prompts a reevaluation in educational and professional settings, where the primary goal is fostering enduring innovative capabilities. A balanced approach is necessary, one tha uses AI tools for immediate creative output while also cultivating and maintaining inherent creative abilities over time.

Secondly, our study echoes the challenges posed by ChatGPT to innovation and diversity in scientific research. The trend toward homogenization of thoughts, evidenced by extensive use of generative AI tools like ChatGPT in creative tasks, poses significant challenges to the innovation and diversity essential in scientific research. As demonstrated by our research, the potential for convergence of ideas may stifle originality and diversity, which are crucial for scientific breakthroughs. This situation calls for strategic policy-making by research institutions, funding agencies, and decision-makers (Bockting et al., 2023; Grimes, Von Krogh, Feuerriegel, Rink, & Gruber, 2023; Hutson & M, 2023). To ensure diversity in scientific thought and innovation, a balance between the efficiencies provided by AI and the encouragement of diverse, original scientific inquiry is needed. This might involve policies that promote varied methodologies and perspectives in research projects, even as AI tools become increasingly integrated into the scientific process.

Finally, the implications of long-term dependency on ChatGPT and the resulting skill degradation warrant attention. The temporary enhancement in creativity by ChatGPT, potentially leading to long-term dependency, highlights concerns about possible skill degradation. This is particularly relevant in educational and professional training environments. An over-reliance on AI for creative tasks could lead to a decline in the development and retention of fundamental creative skills. Therefore, a balanced approach is essential when integrating AI into learning and professional domains. The focus should not be limited to immediate efficiency gains but should also encompass the development and preservation of critical creative skills, ensuring that AI tools are used as supplements to, rather than replacements for, human creativity.

These implications emphasize the need for a nuanced and balanced approach to integrating AI into creative and scientific domains. As we harness the benefits of AI, it's critical to remain cognizant of its limitations and potential impacts on creativity, innovation, skill development, and ethical standards in these fields. This understanding will guide the responsible and effective use of AI to enhance human potential while safeguarding the core values and skills that drive innovation and creativity.

**4.2  Limitations and future directions**

While our experiment offers insightful revelations, it is not without its limitations. To minimize external interferences, we opted for a laboratory setting and conducted two types of creativity tasks to comprehensively assess individual creativity levels. However, this approach also meant a lack of field research, particularly for high-complexity tasks like developing innovative solutions for corporate products.

Moreover, the study's composition, featuring 8 low-complexity AUT tasks and 7 high-complexity product solution tasks, resulted in a substantial collection of 3302 ideas and 427 solutions. Nevertheless, the participant pool was limited to 61 university students, encompassing undergraduates, graduates, and Ph.D. candidates. This raises questions about the sample's representativeness, a limitation we accepted in exchange for ensuring participant retention and high-quality engagement over the crucial one-week period, necessary to foster a dependency on ChatGPT in creative tasks.

Additionally, similar to other researchers, we conducted a follow-up survey, but it was limited to a single instance, one month after the initial experiment. This approach was designed to capture the essence of "sustained innovation" and enhance the external validity of our findings. However, we advocate for future research to consider longer follow-up intervals to more accurately gauge the enduring effects of AI tools like ChatGPT on creativity and innovation.

    Despite these constraints, the core findings of our research offer broad applicability. The fundamental AUT task proved effective in measuring divergent thinking, with robust results observed on both the seventh and thirtieth days. This study lays the groundwork for future exploration into the real-world impacts of ChatGPT in professional settings, particularly its potential to narrow the scope of group innovation and induce ideational convergence. There is a pressing need for further research, employing extended follow-up periods, to unravel the long-term effects of ChatGPT dependency on human creativity. The question of whether such dependency's adverse impacts exacerbate or diminish over time remains a critical avenue for future inquiry.

**Appendix**

**Figure2a Task 1 (AUT) Bar chart of each variable daily**

| Day1 | Day2 | Day3 | Day4 | Day5 | Day6 | Day7 |

**Recognition accuracy - Novelty**

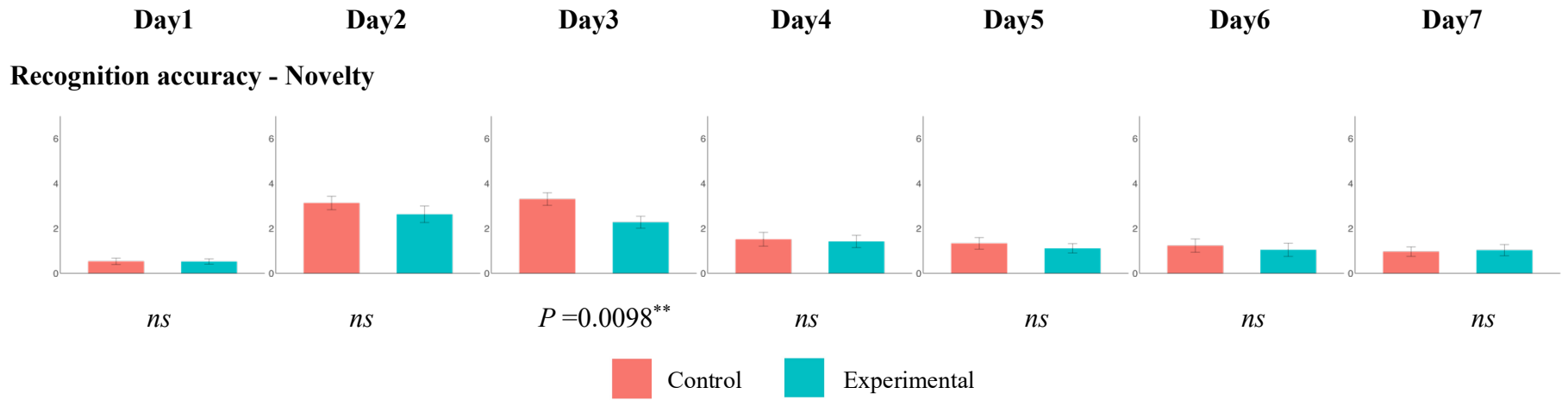

**Recognition accuracy - Usefulness**

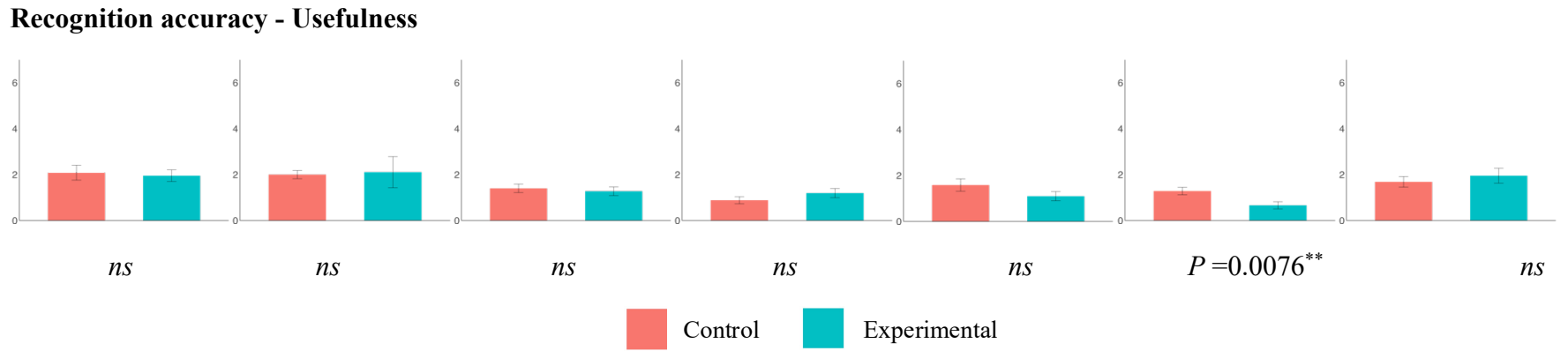

**Figure2b Task 2 Bar chart of each variable daily**

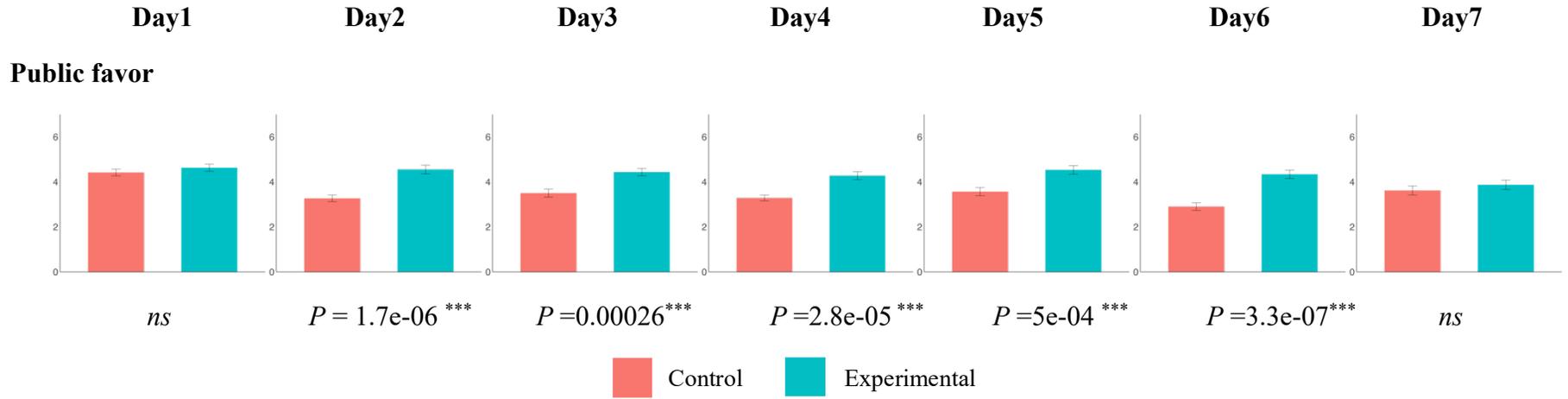

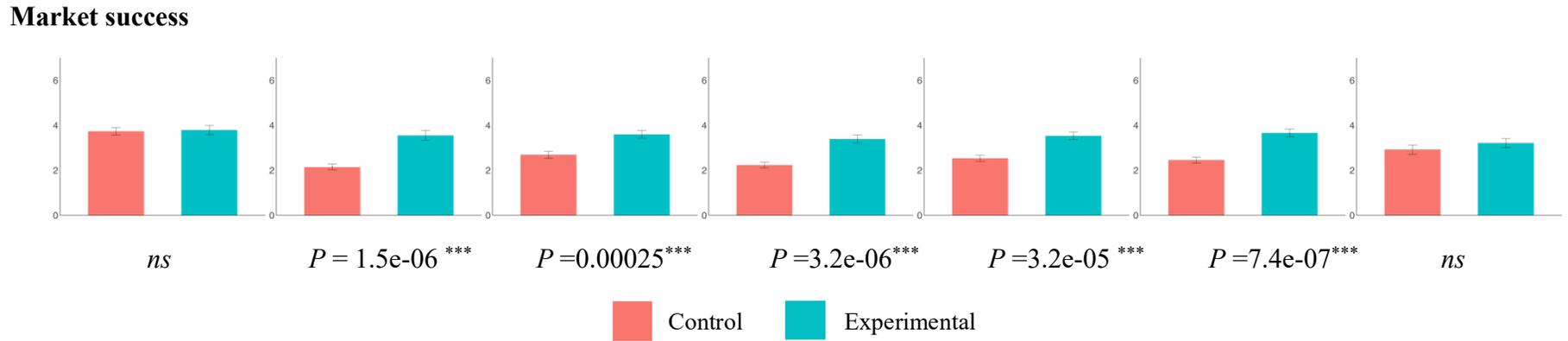

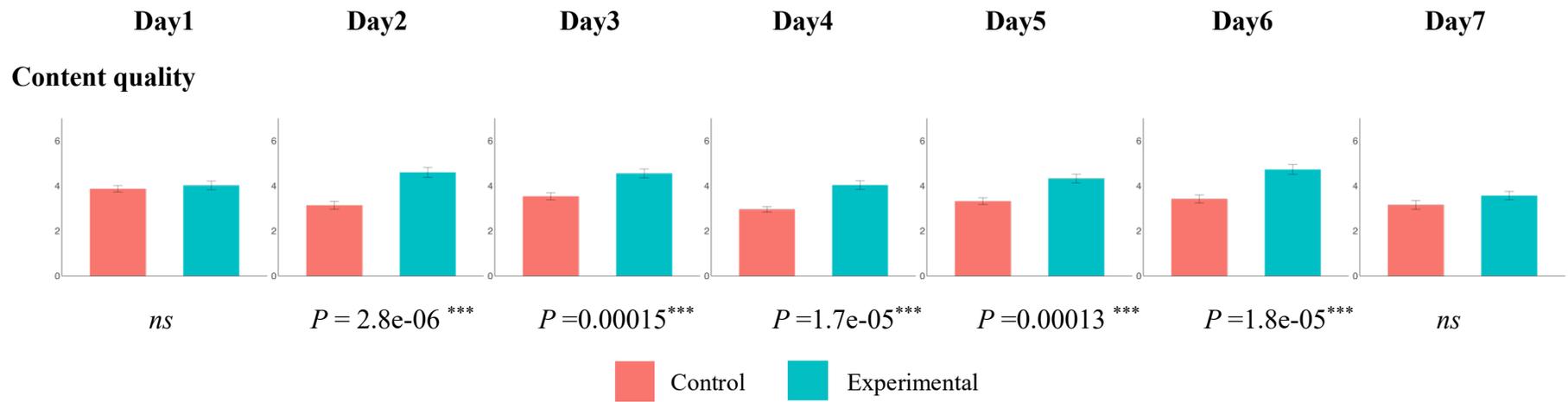